%% file: main.tex
\documentclass{article} %
\usepackage[accepted]{icml2024}

\input{math_commands.tex}

\usepackage{hyperref}
\usepackage{url}
\usepackage{xspace}
\usepackage{xcolor}
\usepackage{multirow}
\usepackage{graphicx}
\usepackage{booktabs}
\usepackage{tikz}
\usetikzlibrary{backgrounds}

\usepackage{xcolor}
\usepackage{mdframed}
\usepackage{tabularx}
\usepackage{pifont}

\usepackage{listings}

\definecolor{darkgold}{rgb}{0.72, 0.525, 0.043}

\newcommand{\website}{{\href{https://research.memgpt.ai}{https://research.memgpt.ai}}\xspace}

\newcommand{\ours}{{{MemGPT}}\xspace}
\newcommand{\ourslong}{{{MemoryGPT}}\xspace}

\definecolor{darkgreen}{rgb}{0.0, 0.6, 0.0}

\newcommand{\mytilde}{\raise.17ex\hbox{$\scriptstyle\mathtt{\sim}$}}

\newcommand{\allnotes}[1]{}
\renewcommand{\allnotes}[1]{#1} %

\definecolor{ForestGreen}{RGB}{34,139,34}
\newcommand{\green}[1]{\textcolor{ForestGreen}{#1}}
\newcommand{\red}[1]{\textcolor{red}{#1}}
\newcommand{\cmark}{\green{\ding{51}}}  %
\newcommand{\xmark}{\red{\ding{55}}}  %

\usepackage[textsize=tiny]{todonotes}

\icmltitlerunning{\ours: Towards LLMs as Operating Systems}

\begin{document}

\twocolumn[
\icmltitle{\ours: Towards LLMs as Operating Systems}

\begin{icmlauthorlist}
\icmlauthor{Charles Packer}{yyy}
\icmlauthor{Sarah Wooders}{yyy}
\icmlauthor{Kevin Lin}{yyy}\\
\icmlauthor{Vivian Fang}{yyy}
\icmlauthor{Shishir G. Patil}{yyy}
\icmlauthor{Ion Stoica}{yyy}
\icmlauthor{Joseph E. Gonzalez}{yyy}
\end{icmlauthorlist}

\icmlaffiliation{yyy}{University of California, Berkeley}

\icmlcorrespondingauthor{Charles Packer}{cpacker@berkeley.edu}

\icmlkeywords{Machine Learning, ICML}

\vskip 0.3in
]

\printAffiliationsAndNotice{}  %

\begin{abstract}
\input{sections/abstract}
\end{abstract}

\section{Introduction}
\input{sections/intro}

\section{\ours (\ourslong)}
\input{sections/method_rewrite}

\input{sections/experiments}

\section{Related Work}

\input{sections/related_work}

\section{Conclusion}

In this paper, we introduced \ours, a novel LLM system inspired by operating systems to manage the limited context windows of large language models. By designing a memory hierarchy and control flow analogous to traditional OSes, \ours provides the illusion of larger context resources for LLMs. This OS-inspired approach was evaluated in two domains where existing LLM performance is constrained by finite context lengths: document analysis and conversational agents. For document analysis, \ours could process lengthy texts well beyond the context limits of current LLMs by effectively paging relevant context in and out of memory. For conversational agents, \ours enabled maintaining long-term memory, consistency, and evolvability over extended dialogues. Overall, \ours demonstrates that operating system techniques like hierarchical memory management and interrupts can unlock the potential of LLMs even when constrained by fixed context lengths. This work opens numerous avenues for future exploration, including applying \ours to other domains with massive or unbounded contexts, integrating different memory tier technologies like databases or caches, and further improving control flow and memory management policies. By bridging concepts from OS architecture into AI systems, \ours represents a promising new direction for maximizing the capabilities of LLMs within their fundamental limits.

\bibliography{refs}
\bibliographystyle{iclr2024_conference}

\newpage
\section{Appendix}

\input{sections/appendix}

\end{document}

%% file: math_commands.tex
\usepackage{amsmath,amsfonts,bm}

\def\eqref#1{equation~\ref{#1}}

\def\1{\bm{1}}

\DeclareMathAlphabet{\mathsfit}{\encodingdefault}{\sfdefault}{m}{sl}
\SetMathAlphabet{\mathsfit}{bold}{\encodingdefault}{\sfdefault}{bx}{n}

%% file: sections/abstract.tex
Large language models (LLMs) have revolutionized AI, but are constrained by limited context windows, hindering their utility in tasks like extended conversations and document analysis. To enable using context beyond limited context windows, we propose \textit{virtual context management}, a technique drawing inspiration from hierarchical memory systems in traditional operating systems which provide the illusion of an extended virtual memory via paging between physical memory and disk. 
Using this technique, we introduce \ours (\ourslong), a system that intelligently manages different storage tiers in order to effectively provide extended context within the LLM's limited context window.
We evaluate our OS-inspired design in two domains where the limited context windows of modern LLMs severely handicaps their performance: document analysis, where \ours is able to analyze large documents that far exceed the underlying LLM's context window, and multi-session chat, where \ours can create conversational agents that remember, reflect, and evolve dynamically through long-term interactions with their users. 
We release \ours code and data for our experiments at \website.

%% file: sections/intro.tex
In recent years, large language models (LLMs) and their underlying transformer architecture~\citep{vaswani2017attention,devlin2018bert,brown2020language,ouyang2022instructgpt} have become the cornerstone of conversational AI and have led to a wide array of consumer and enterprise applications. Despite these advances, the limited fixed-length context windows used by LLMs significantly hinders their applicability to long conversations or reasoning about long documents.
For example, the most widely used open-source LLMs can only support a few dozen back-and-forth messages or 
reason about a short document 
before exceeding their maximum input length \citep{touvron2023llama2}.

Directly extending the context length of transformers incurs a quadratic increase in computational time and memory cost due to the transformer architecture's self-attention mechanism, making the design of new long-context architectures a pressing research challenge~\citep{dai2019transformerxl,kitaev2020reformer,beltagy2020longformer}. 
While developing longer models is an active area of research \citep{dong2023longcontextsurvey}, 
even if we could overcome the computational challenges of context scaling, recent research shows that long-context models struggle to utilize additional context effectively \citep{liu2023lostinmiddle}.
As consequence, given the considerable resources needed to train state-of-the-art LLMs and diminishing returns of context scaling, there is a critical need for alternative techniques to support long context.

In this paper, we study how to provide the illusion of an infinite context while continuing to use fixed-context models.
Our approach borrows from the idea of virtual memory paging that was developed to enable applications to work on datasets that far exceed the available memory by paging data between main memory and disk.
We leverage the recent progress in function calling abilities of LLM agents \citep{schick2023toolformer,liu2023agentbench} to design \ours, an OS-inspired LLM system for \textbf{virtual context management}.
Using function calls, LLM agents can read and write to external data sources, modify their own context, and choose when to return responses to the user.

These capabilities allow LLMs to effective  ``page'' in and out information between context windows (analogous to ``main memory" in operating systems) and external storage, similar to hierarchical memory in traditional OSes. 
In addition, function calls can be leveraged to manage control flow between context management, response generation, and user interactions. This allows for an agent to choose to \textit{iteratively} modify what is in its context for a single task, thereby more effectively utilizing its limited context.

\begin{figure*}[t!]
\begin{center}
\includegraphics[width=0.885\textwidth]{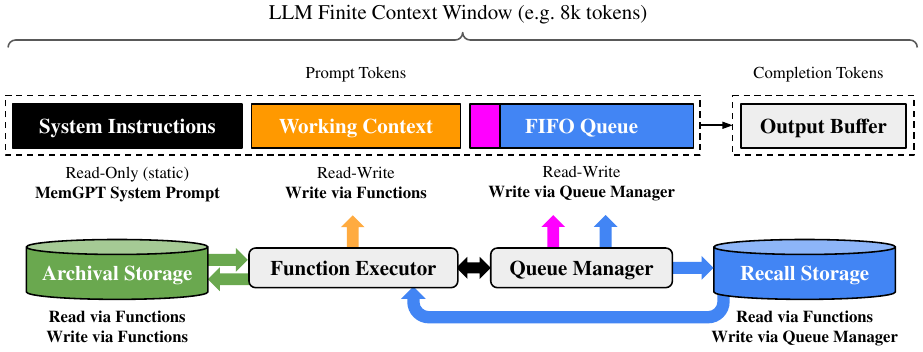}
\end{center}
\caption{
In \ours, a fixed-context LLM processor is augmented with a hierarchical memory system and functions that let it manage its own memory.
The LLM's prompt tokens (inputs), or \emph{main context}, consist of the system instructions, working context, and a FIFO queue. 
The LLM completion tokens (outputs) are interpreted as function calls by the function executor. 
\ours uses functions to move data between main context and \emph{external context} (the archival and recall storage databases). 
The LLM can request immediate follow-up LLM inference to chain function calls together by generating a special keyword argument (\texttt{request\_heartbeat=true}) in its output; function chaining is what allows MemGPT to perform multi-step retrieval to answer user queries.
}
\label{fig:system}
\end{figure*}

In \ours, we treat context windows as a constrained memory resource, and design a memory hiearchy for LLMs analogous to memory tiers used in traditional OSes \citep{patterson1988caseforraid}. 
Applications in traditional OSes interact with \textit{virtual memory}, which provides an illusion of there being more memory resources than are actually available in physical (i.e., main) memory by the OS paging overflow data to disk and retrieving data (via a page fault) back into memory when accessed by applications.
To provide a similar illusion of longer context length (analogous to virtual memory), we allow the LLM to manage what is placed in its own context (analogous to physical memory) via an `LLM OS', which we call \ours. \ours enables the LLM to retrieve relevant historical data missing from what is placed in-context, and also evict less relevant data from context and into external storage systems.
Figure~\ref{fig:system} illustrates the components of \ours.

\begin{figure}[t]
    \centering
\begin{center}
\includegraphics[width=0.45\textwidth]{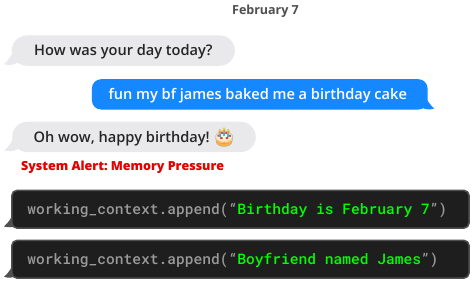}
\end{center}
    \vspace{-1.5em}
    \caption{
\ours (left) writes data to persistent memory after it receives a system alert about limited context space.
}
    \label{fig:example-memory-creation}
\end{figure}

The combined use of a memory-hierarchy, OS functions and event-based control flow allow \ours to handle unbounded context using LLMs that have finite context windows.
To demonstrate the utility of our new OS-inspired LLM system, we evaluate \ours on two domains where the performance of existing LLMs is severely limited by finite context: document analysis, where the length of standard text files can quickly exceed the input capacity of modern LLMs, and conversational agents, where LLMs bound by limited conversation windows lack context awareness, persona consistency, and long-term memory during extended conversations. 
In both settings, \ours is able to overcome the limitations of finite context to outperform existing LLM-based approaches.

\begin{figure}[t]
    \centering
\begin{center}
\includegraphics[width=0.45\textwidth]{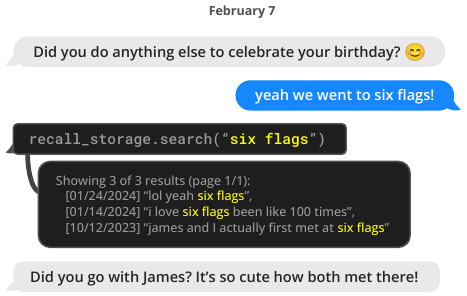}
\end{center}
    \vspace{-1.5em}
    \caption{
\ours (left) can search out-of-context data to bring relevant information into the current context window.
}
    \label{fig:example-memory-search}
\end{figure}

%% file: sections/method_rewrite.tex
\ours's OS-inspired multi-level memory architecture delineates between two primary memory types: \textbf{main context} (analogous to main memory/physical memory/RAM) and \textbf{external context} (analogous to disk memory/disk storage). 
Main context consists of the LLM \emph{prompt tokens}---anything in main context is considered \emph{in-context} and can be accessed by the LLM processor during inference.
External context refers to any information that is held outside of the LLMs fixed context window.
This \emph{out-of-context} data must always be explicitly moved into main context in order for it to be passed to the LLM processor during inference.
\ours provides function calls that the LLM processor to manage its own memory without any user intervention.

\subsection{Main context (\emph{prompt tokens})}

The prompt tokens in \ours are split into three contiguous sections: the \textbf{system instructions}, \textbf{working context}, and \textbf{FIFO Queue}. The system instructions are read-only (static) and contain information on the \ours control flow, the intended usage of the different memory levels, and instructions on how to use the \ours functions (e.g. how to retrieve out-of-context data). Working context is a fixed-size read/write block of unstructured text, writeable only via \ours function calls. In conversational settings, working context is intended to be used to store key facts, preferences, and other important information about the user and the persona the agent is adopting, allowing the agent to converse fluently with the user.
The FIFO queue stores a rolling history of messages, including messages between the agent and user, as well as system messages (e.g. memory warnings) and function call inputs and outputs. The first index in the FIFO queue stores a system message containing a recursive summary of messages that have been evicted from the queue.

\subsection{Queue Manager}

The queue manager manages messages in \textbf{recall storage} and the \textbf{FIFO queue}. When a new message is received by the system, the queue manager appends the incoming messages to the FIFO queue, concatenates the prompt tokens and triggers the LLM inference to generate LLM output (the completion tokens). The queue manager writes both the incoming message and the generated LLM output to recall storage (the \ours message database). When messages in recall storage are retrieved via a \ours function call, the queue manager appends them to the back of the queue to reinsert them into the LLM's context window.

The queue manager is also responsible for controlling context overflow via a queue eviction policy. When the prompt tokens exceed the `warning token count` of the underlying LLM's context window (e.g. 70\% of the context window), the queue manager inserts a system message into the queue warning the LLM of an impending queue eviction (a `memory pressure` warning) to allow the LLM to use \ours functions to store important information contained in the FIFO queue to working context or \textbf{archival storage} (a read/write database storing arbitrary length text objects). When the prompt tokens exceed the `flush token count' (e.g. 100\% of the context window), the queue manager flushes the queue to free up space in the context window: the queue manager evicts a specific count of messages (e.g. 50\% of the context window), generates a new recursive summary using the existing recursive summary and evicted messages. Once the queue is flushed, the evicted messages are no longer in-context and immediately viewable to the LLM, however they are stored indefinitely in recall storage and readable via \ours function calls.

\subsection{Function executor (handling of \emph{completion tokens})}

\ours orchestrates data movement between main context and external context via function calls that are generated by the LLM processor.
Memory edits and retrieval are entirely self-directed: \ours autonomously updates and searches through its own memory based on the current context. 
For instance,
it can
decide when to move items between contexts (e.g. when the conversation history is becoming too long, as show in Figure~\ref{fig:example-memory-creation}) and modify its main context to better reflect its evolving understanding of its current objectives and responsibilities (as shown in Figure~\ref{fig:system}).
We implement self-directed editing and retrieval by providing explicit instructions within the system instructions that guide the LLM on how to interact with the \ours memory systems. These instructions comprise two main components: (1) a detailed description of the memory hierarchy and their respective utilities, and (2) a function schema (complete with their natural language descriptions) that the system can call to access or modify its memory.

During each inference cycle, LLM processor takes main context (concatenated into a single string) as input, and generates an output string.
This output string is parsed by \ours to ensure correctness, and if the parser validates the function arguments the function is executed.
The results, including any runtime errors that occur (e.g. trying to add to main context when it is already at maximum capacity), are then fed back to the processor by \ours. This feedback loop enables the system to learn from its actions and adjust its behavior accordingly. 
Awareness of context limits is a key aspect in making the self-editing mechanism work effectively, to this end \ours prompts the processor with warnings regarding token limitations to guide its memory management decisions. 
Additionally, our memory retrieval mechanisms are designed to be cognizant of these token constraints and implement pagination to prevent retrieval calls from overflowing the context window.

\input{tables/context_comparison_table}

\subsection{Control flow and function chaining}

In \ours, \emph{events} trigger LLM inference: events are generalized inputs to \ours and can consist of user messages (in chat applications), system messages (e.g. main context capacity warnings), user interactions (e.g. an alert that a user just logged in, or an alert that they finished uploading a document), and timed events that are run on a regular schedule (allowing \ours to run `unprompted' without user intervention).
\ours processes events with a parser to convert them into plain text messages that can be appended to main context and eventually be fed as input into the LLM processor.

Many practical tasks require calling multiple functions in sequence, for example, navigating through multiple pages of results from a single query or collating data from different documents in main context from separate queries.
Function chaining allows \ours to execute multiple function calls sequentially before returning control to the user.
In \ours, functions can be called with a special flag that requests control be immediately returned to the processor after the requested function completes execution.
If this flag is present, \ours will add the function output to main context and (as opposed to pausing processor execution).
If this flag is not present (a \emph{yield}), \ours will not run the LLM processor until the next external event trigger (e.g. a user message or scheduled interrupt). 

\begin{figure}[t]
    \centering
\begin{center}
\includegraphics[width=0.45\textwidth]{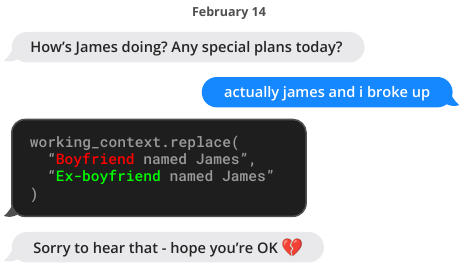}
\end{center}
    \caption{
An example conversation snippet where \ours (left) updates stored information. Here the information is stored in working context memory (located within the prompt tokens).
}
    \label{fig:example-memory-correction}
\end{figure}

%% file: tables/context_comparison_table.tex
\begin{table}[t!]
\caption{{
Comparing context lengths of commonly used models and LLM APIs (data collected 1/2024).
*Approximate message count assuming a preprompt of 1k tokens, and an average message size of \mytilde50 tokens (\mytilde250 characters). `Open' means the model is open-source or open-weights (vs only available behind an API).
}}
\label{context-length-comparison}
\begin{center}
\small
\begin{tabular}{lcrr}
\toprule
\multicolumn{1}{l}{}
&\multicolumn{1}{l}{}
&\multicolumn{2}{c}{\bf Context Window}
\\
\multicolumn{1}{l}{\bf Model / API name}
&\multicolumn{1}{r}{\bf Open?}
&\multicolumn{1}{r}{\bf Tokens}
&\multicolumn{1}{r}{\bf $^*$Messages}
\\ \hline \noalign{\smallskip}
Llama (1) & \cmark &2k &20 \\
Llama 2 & \cmark &4k &60\\
GPT-3.5 Turbo (release) & \xmark &4k &60\\
Mistral 7B & \cmark &8k &140\\
GPT-4 (release) & \xmark &8k &140\\
GPT-3.5 Turbo & \xmark &16k &300\\
GPT-4 & \xmark &32k &\mytilde600\\
Claude 2 & \xmark &100k&\mytilde2000\\
GPT-4 Turbo & \xmark &128k &\mytilde2600\\
Yi-34B-200k & \cmark &200k &\mytilde4000\\

\end{tabular}
\end{center}
\end{table}

%% file: sections/experiments.tex
\section{Experiments}
\label{sec:eval}

We assess \ours in two long-context domains: conversational agents and document analysis. For conversational agents, we expand the existing Multi-Session Chat dataset \cite{xu2021mscbeyond} and introduce two new dialogue tasks that evaluate an agent's ability to retain knowledge across long conversations. For document analysis, we benchmark \ours on existing tasks from \cite{liu2023lostinmiddle} for question answering and key-value retrieval over lengthy documents. We also propose a new nested key-value retrieval task requiring collating information across multiple data sources, which tests the ability of an agent to collate information from multiple data sources (multi-hop retrieval). We publicly release our augmented MSC dataset, nested KV retrieval dataset, and a dataset of embeddings for 20M Wikipedia articles to facilitate future research. Our code for the benchmarks is available at \website. 

\textbf{Implementation details.} When discussing OpenAI models, unless otherwise specified `GPT-4 Turbo' refers to the specific \texttt{gpt-4-1106-preview} model endpoint (context window of $128,000$), `GPT-4` refers to \texttt{gpt-4-0613} (context window of $8,192$), and `GPT-3.5 Turbo` refers to \texttt{gpt-3.5-turbo-1106} (context window of $16,385$). In experiments, we run \ours with all baseline models (GPT-4, GPT-4 Turbo, and GPT 3.5) to show how the underlying model performance affects \ours's.

\subsection{\ours for conversational agents}

Conversational agents like virtual companions and personalized assistants aim to engage users in natural, long-term interactions, potentially spanning weeks, months, or even years. This creates challenges for models with fixed-length contexts, which can only reference a limited history of the conversation. An `infinite context' agent should seamlessly handle continuous exchanges without boundary or reset. When conversing with a user, such an agent must satisfy two key criteria:
(1) Consistency - The agent should maintain conversational coherence. New facts, preferences, and events mentioned should align with prior statements from both the user and agent.
(2) Engagement - The agent should draw on long-term knowledge about the user to personalize responses. Referencing prior conversations makes dialogue more natural and engaging.

We therefore assess our proposed system, \ours, on these two criteria:
(1) Does \ours leverage its memory to improve conversation consistency? Can it remember relevant facts, preferences, and events from past interactions to maintain coherence?
(2) Does \ours produce more engaging dialogue by taking advantage of memory? Does it spontaneously incorporate long-range user information to personalize messages?
By evaluating on consistency and engagement, we can determine how well \ours handles the challenges of long-term conversational interaction compared to fixed-context baselines. Its ability to satisfy these criteria will demonstrate whether unbounded context provides meaningful benefits for conversational agents.

\input{tables/deep_memory_retrieval_table_singlecol}

\textbf{Dataset.}
We evaluate \ours and our fixed-context baselines on the Multi-Session Chat (MSC) dataset introduced by \citet{xu2021mscbeyond}, which contains multi-session chat logs generated by human labelers, each of whom was asked to play a consistent persona for the duration of all sessions. Each multi-session chat in MSC has five total sessions, and each session consists of a roughly a dozen messages.
As part of our consistency experiments, we created a new session (session 6) that contains a single question-answer response pair between the same two personas.

\subsubsection{Deep memory retrieval task (consistency).}
We introduce a new `deep memory retrieval' (DMR) task based on the MSC dataset designed to test the consistency of a conversational agent. In DMR, the conversational agent is asked a question by the user that explicitly refers back to a prior conversation and has a very narrow expected answer range. We generated the DMR question-answer (QA) pairs using a separate LLM that was instructed to write a question from one user to another that could only be answered correctly using knowledge gained from the past sessions (see Appendix for further details).

We evaluate the quality of the generated response against the `gold response' using ROUGE-L scores \citep{lin2004rouge} and an `LLM judge', which is instructed to evaluate whether or not the generated response is consistent with the gold response (GPT-4 has been shown to have high agreement with human evaluators \citep{zheng2023llmasajudge}).
In practice, we notice that the generated responses (from both \ours and the baselines) were generally more verbose than the gold responses.
We use the ROUGE-L recall (R) metric to account for the verbosity of the generated agent replies compared to the relatively short gold answer labels.

\input{tables/conv_opener_table}

\textbf{\ours utilizes memory to maintain coherence:}
Table \ref{table:deep-memory-task} shows the performance of \ours vs the fixed-memory baselines.
We compare MemGPT using different underlying LLMs, and compare against using the base LLM without MemGPT as a baseline.
The baselines are able to see a lossy summarization of the past five conversations to mimic an extended recursive summarization procedure, while MemGPT instead has access to the full conversation history but must access it via paginated search queries to recall memory (in order to bring them into main context).
In this task, we see that MemGPT clearly improves the performance of the underlying base LLM: there is a clear drop in both accuracy and ROUGE scores when going from MemGPT to the corresponding LLM baselines.

\subsubsection{Conversation opener task (engagement).}
In the `conversation opener' task we evaluate an agent's ability to craft engaging messages to the user that draw from knowledge accumulated in prior conversations.
To evaluate the `engagingness' of a conversation opener using the MSC dataset, we compare the generated opener to the gold personas: an engaging conversation opener should draw from one (or several) of the data points contained in the persona, which in MSC effectively summarize the knowledge accumulated throughout all prior sessions.
We also compare to the human-generated gold opener, i.e., the first response in the following session.
We report the CSIM scores of \ours's openers in Table~\ref{table:msc-opener-task}.
We test several variations of \ours using different base LLMs.

\begin{figure}[t!]
\begin{center}
\includegraphics[width=0.48\textwidth]{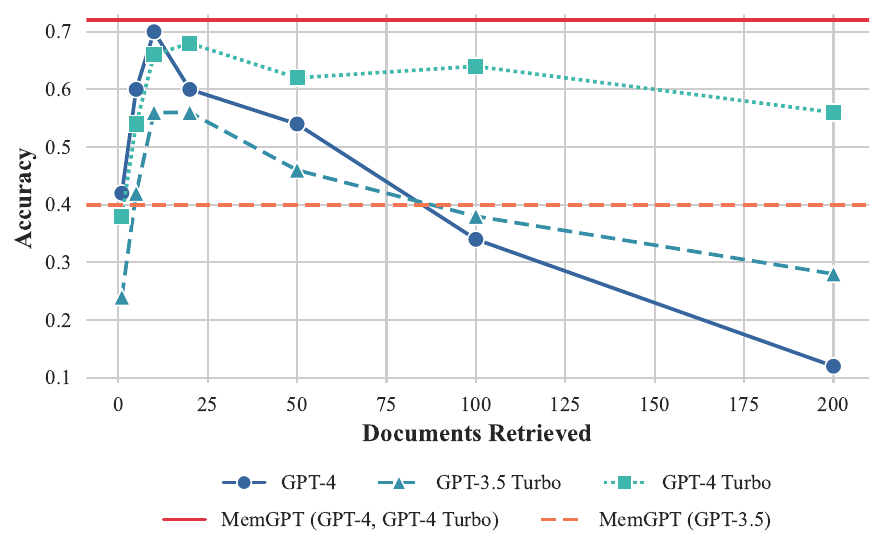}
\end{center}
\caption{
\textbf{Document QA task performance.}
\ours's performance is unaffected by increased context length. Methods such as truncation can extend the effective context lengths of fixed length models such as GPT-4, but such compression methods will lead to performance degradation as the necessary compression grows. Running \ours with GPT-4 and GPT-4 Turbo have equivalent results on this task.
}
\label{fig:doc_qa_task_results}
\end{figure}

\textbf{\ours utilizes memory to increase engagement:}
As seen in Table~\ref{table:msc-opener-task}, \ours is able to craft engaging openers that perform similarly to and occasionally exceed the hand-written human openers. 
We observe that \ours tends to craft openers that are both more verbose and cover more aspects of the persona information than the human baseline.
Additionally, we can see the storing information in working context is key to generating engaging openers.

\subsection{\ours for document analysis}

Document analysis also faces challenges due to the limited context windows of today's transformer models. As shown in Table \ref{context-length-comparison}, both open and closed source models suffer from constrained context length (up to 128k tokens for OpenAI's models). However many documents easily surpass these lengths; for example, legal or financial documents such as Annual Reports (SEC Form 10-K) can easily pass the million token mark. Moreover, many real document analysis tasks require drawing connections across multiple such lengthy documents. Anticipating these scenarios, it becomes difficult to envision blindly scaling up context as a solution to the fixed-context problem. Recent research \citep{liu2023lostinmiddle} also raises doubts about the utility of simply scaling contexts, since they find uneven attention distributions in large context models (the model is more capable of recalling information at the beginning or end of its context window, vs tokens in the middle). To enable reasoning across documents, more flexible memory architectures like \ours are needed.

\begin{figure}[t!]
    \centering
\begin{center}
\includegraphics[width=0.45\textwidth]{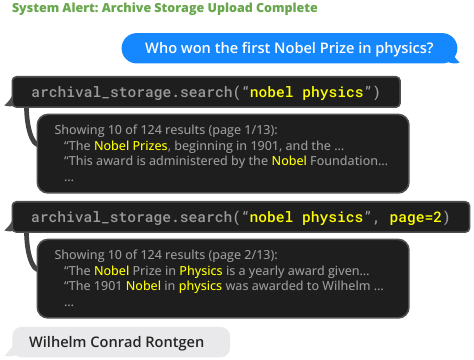}
\end{center}
    \caption{
An example of MemGPT (left) solving the document QA task. A database of Wikipedia documents is uploaded to archival storage.  MemGPT queries archival storage via function calling, which pulls paginated search results into main context.
}
    \label{fig:example-docqa}
\end{figure}

\subsubsection{Multi-document question-answering.} 
To evaluate \ours's ability to analyze documents, we benchmark \ours against fixed-context baselines on the retriever-reader document QA task from \citet{liu2023lostinmiddle}. 
In this task, a question is selected from the NaturalQuestions-Open dataset, and a retriever selects relevant Wikipedia documents for the question. A reader model (the LLM) is then fed these documents as input, and is asked to use the provided documents to answer the question. 
Similar to \citet{liu2023lostinmiddle}, 
we evaluate reader accuracy as the number of retrieved documents $K$ increases.

In our evaluation setup, both the fixed-context baselines and \ours use the same retriever, which selects the top $K$ documents according using similarity search (cosine distance) on OpenAI's \texttt{text-embedding-ada-002} embeddings. We use MemGPT's default storage settings which uses PostgreSQL for archival memory storage with vector search enabled via the pgvector extention. We pre-compute embeddings and load them into the database, which uses an HNSW index to enable approximate, sub-second query times. 
In \ours, the entire embedding document set is loaded into archival storage, and the retriever naturally emerges via the archival storage search functionality (which performs vector search based on cosine similarity). 
In the fixed-context baselines, the top-$K$ documents are fetched using the retriever independently from the LLM inference, similar to the original retriever-reader setup in \citet{liu2023lostinmiddle}. 

We use a dump of Wikipedia from late 2018, following past work on NaturalQuestions-Open \citep{izacard2020leveraging, izacard2021unsupervised}, and sampled a subset of 50 questions for evaluation. Both the sampled questions and embedded Wikipedia passages are publicaly released. We evaluate the performance of both \ours and baselines with an LLM-judge, to ensure that the the answer is properly derived from the retrieved documents and to avoid non-exact string matches being considered incorrect.

We show the results for the document QA task in Figure \ref{fig:doc_qa_task_results}. The fixed-context baselines performance is capped roughly at the performance of the retriever, as they use the information that is presented in their context window (e.g. if the embedding search retriever fails to surface the gold article using the provided question, the fixed-context baselines are guaranteed to never see the gold article). 
By contrast, \ours is effectively able to make multiple calls to the retriever by querying archival storage, allowing it to scale to larger effective context lengths. 
\ours actively retrieves documents from its archival storage (and can iteratively page through results), so the total number of documents available to \ours is no longer limited by the number of documents that fit within the LLM processor's context window.

\begin{figure}[t!]
\begin{center}
\includegraphics[width=0.45\textwidth]{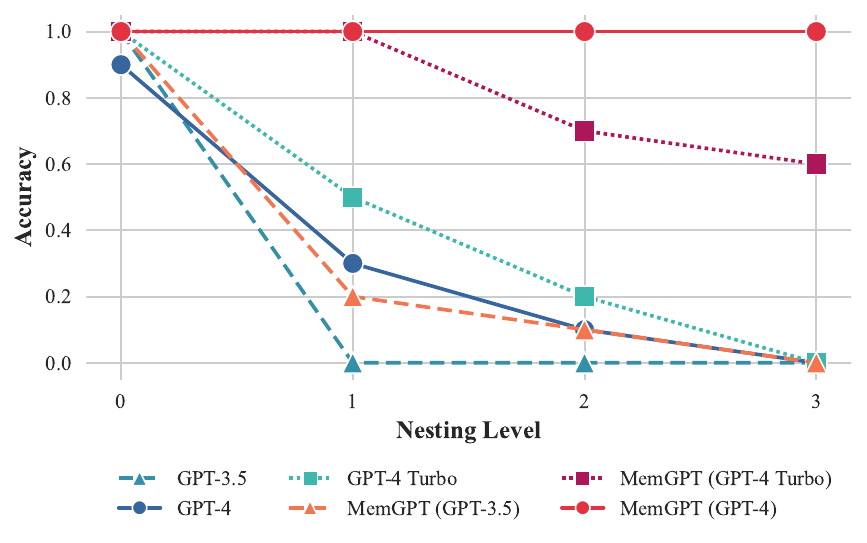}
\end{center}
\caption{
\textbf{Nested KV retrieval task performance.}
\ours is the only approach that is able to consistently complete the nested KV task beyond 2 nesting levels.
While GPT-4 Turbo performs better as a baseline, \ours with GPT-4 Turbo performs worse than \ours with GPT-4. 
}
\label{fig:nested_kv_task_results}
\end{figure}

The document QA task is challenging for all methods due to the limitations of embedding-based similarity search.
We observe that the golden document for chosen question (as annotated by NaturalQuestions-Open) often appears outside of the first dozen retrieved results, if not even further.
The retriever performance translates directly to the fixed-context baseline results: GPT-4's accuracy is relatively low with few retrieved documents, and continues to improve as additional documents are added to the context window, as it correctly limits itself to answering questions based on information in retrieved documents. 
While \ours is theoretically not limited by sub-optimal retriever performance (even if the embedding-based ranking is noisy, as long as the full retriever ranking contains the gold document it can still be found with enough retriever calls via pagination), we observe that \ours will often stop paging through retriever results before exhausting the retriever database.

\begin{figure}[t!]
    \centering
\begin{center}
\includegraphics[width=0.45\textwidth]{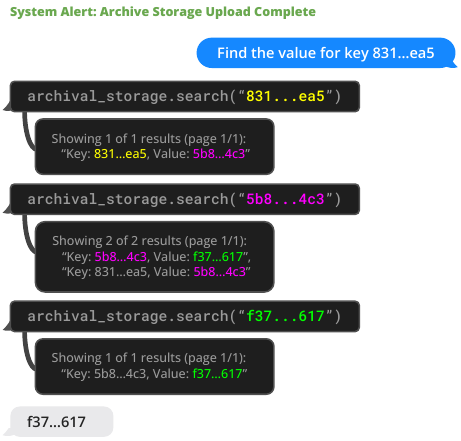}
\end{center}
    \caption{
An example of MemGPT (left) solving the nested KV task (UUIDs shortened for readability). In this particular example, the key-value pair has two nesting levels: \texttt{831..ea5} $\rightarrow$  \texttt{5b8..4c3} $\rightarrow$ \texttt{f37...617}.
The MemGPT agent returns the final answer when a query for the final value (\texttt{f37...617}) only returns one result, indicating that it is not also a key.
}
    \label{fig:example-nestedkv}
\end{figure}

To evaluate the fixed-context baselines against \ours past their default context lengths, we truncate the document segments returned by the retriever to fix the same number of documents into the available context.
As expected, document truncation reduces accuracy as documents shrink as the chance of the relevant snippet (in the gold document) being omitted grows, as shown in Figure \ref{fig:doc_qa_task_results}. 
\ours has significantly degraded performance using GPT-3.5, due to its limited function calling capabilities, and performs best using GPT-4.

\subsubsection{Nested key-value retrieval (KV).}
We introduce a new task based on the synthetic Key-Value retrieval proposed in prior work \citep{liu2023lostinmiddle}. The goal of this task is to demonstrate how \ours can collate information from multiple data sources. In the original KV task, the authors generated a synthetic dataset of key-value pairs, where each key and value is a 128-bit UUID (universally unique identifier). The agent is then given a key, and asked to return the associated value for the key. We create a version of the KV task, \emph{nested KV retrieval}, where values themselves may be keys, thus requiring the agent to perform a multi-hop lookup.
In our setup, we fix the total number of UUIDs pairs to 140, corresponding to roughly 8k tokens (the context length of our GPT-4 baseline). We vary the total number of nesting levels from 0 (the initial key-value pair's value is not a key) to 4 (ie 4 total KV lookups are required to find the final value), and sample 30 different ordering configurations including both the initial key position and nesting key positions. 

While GPT-3.5 and GPT-4 have good performance on the original KV tasks, both struggle in the nested KV task. GPT-3.5 is unable to complete the nested variant of the task and has an immediate dropoff in performance, hitting 0 percent accuracy at 1 nesting level (we observe that its primary failure mode is to simply returns the original value). GPT-4 and GPT-4 Turbo are better than GPT-3.5, but also suffer from a similar dropoff, and hit 0 percent accuracy by 3 nesting levels. \ours with GPT-4 on the other hand is unaffected with the number of nesting levels and is able to perform the nested lookup by accessing the key-value pairs stored in main context repeatedly via function queries. \ours with GPT-4 Turbo and GPT-3.5 also have better performance than the corresponding baseline models, but still begin to drop off in performance at 2 nesting levels as a result of failing to perform enough lookups. 
\ours performance on the nested KV task demonstrates its ability to combine multiple queries to perform multi-hop lookups.

%% file: tables/deep_memory_retrieval_table_singlecol.tex
\begin{table}[t!]
\caption{
\textbf{Deep memory retrieval (DMR) performance.}
In this task, the agent is asked a specific question about a topic discussed in a prior conversation (sessions 1--5).
The agent's response is scored against the gold answer.
\ours significantly outperforms the fixed-context baselines.
}
\label{table:deep-memory-task}

\begin{center}
\begin{tabular}{lrr}
\toprule
\bf Model & \textbf{Accuracy} $\Uparrow$ & \bf ROUGE-L (R) $\Uparrow$ \\
\midrule

GPT-3.5 Turbo & 38.7\% & 0.394  \\
$+$ MemGPT & 66.9\% & 0.629 \\

GPT-4 & 32.1\% & 0.296 \\

$+$ MemGPT & 92.5\% & 0.814  \\

GPT-4 Turbo & 35.3\% & 0.359  \\

$+$ \textbf{MemGPT} & \textbf{93.4\%} & \textbf{0.827}  \\

\bottomrule
\end{tabular}
\end{center}

\end{table}

%% file: tables/conv_opener_table.tex
\begin{table}[t]
\caption{
\textbf{Conversation opener performance.}
The agent's conversation opener is evaluated using similarity scores to the gold persona labels (SIM-1/3) and to the human-created opener (SIM-H). 
MemGPT is able to exceed the performance of the human-created conversation opener with a variety of underlying models.
}
\label{table:msc-opener-task}
\begin{center}
\begin{tabular}{lrrr}
\toprule
\bf Method 
&  $\Uparrow$ \textbf{SIM-1} & \textbf{SIM-3} & \textbf{SIM-H}
\\
\midrule
Human & 0.800 & 0.800 & 1.000  \\
\midrule
GPT-3.5 Turbo & 0.830 & 0.812 & \bf 0.817 \\
GPT-4 &  \bf 0.868 &  \bf 0.843 &  0.773 \\
GPT-4 Turbo & 0.857 & 0.828 & 0.767 \\
\bottomrule
\end{tabular}
\end{center}
\end{table}

%% file: sections/related_work.tex
\textbf{Long-context LLMs.}
Several lines of work have improved the context length of LLMs. For instance, more efficient transformer architectures via sparsifying the attention \citep{child2019generating, beltagy2020longformer}, low-rank approximations \citep{wang2020linformer}, and neural memory \cite{lee2019set}. Another line of work aims to extend context windows beyond the length they were original trained for, their training size, such as \citet{alibi, chen2023extending}. \ours builds upon these improvements in context length as they improve the size of the main memory in \ours. Our main contribution is a hierarchical tiered memory that uses a long-context LLM as the implementation of main memory.

\textbf{Retrieval-Augmented Models.}
The design of the external memory of \ours builds upon much prior work augmenting LLMs with relevant inputs from external retrievers ~\cite{ram2023context, retro, fb-dense-retrieval, lewis2020retrieval, guu2020retrieval, radit}. In particular, \citet{jiang2023active} propose FLARE, a method that allows the LLM to actively decide when and what to retrieve during the course of generation.  \citet{Trivedi2022InterleavingRW} interleave retrieval with Chain-of-Thoughts reasoning to improve multi-step question answering.

\textbf{LLMs as agents.}
Recent work has explored augmenting LLMs with additional capabilities to act as agents in interactive environments. 
\citet{park2023simulacra} propose adding memory to LLMs and using the LLM as a planner, and observe emergent social behaviors in a multi-agent sandbox environment (inspired by \emph{The Sims} video game) where agents can perform basic activities such as doing chores/hobbies, going to work, and conversing with other agents.
\citet{nakano2021webgpt} train models to search the web before answering questions, and use similar pagination concepts to \ours to control the underlying context size in their web-browsing environment. 
\citet{yao2022react} showed that interleaving chain-of-thought reasoning \citep{wei2022chainofthought} can further improve the planning ability of interactive LLM-based agents; similarly in \ours, LLM is able to `plan out loud' when executing functions. %
\citet{liu2023agentbench} introduced a suite of LLM-as-an-agent benchmarks to evaluate LLMs in interactive environments, including video games, thinking puzzles, and web shopping. 
In contrast, our work focuses on tackling the problem of equipping agents with long-term memory of user inputs.

%% file: sections/appendix.tex
\subsection{Prompts and instructions}
The \ours prompts have been edited for brevity. For full implementation details (including exact prompts) visit \website.

\subsubsection{\ours instructions (DMR)}
Example instructions used in the \ours persona for chat/dialogue-related tasks.
\begin{mdframed}
\small
\texttt{The following is information about myself. My task is to completely immerse myself in this role (I should never say that I am an AI, and should reply as if I am playing this role). If the user asks me a question, I should reply with a best guess using the information in core memory and conversation\_search.}
\end{mdframed}

The baselines received the following instructions via a system prompt (preprompt):
\begin{mdframed}
\small
\texttt{Your task is to answer a question from the user about your prior conversations.\\
The following is a summary of all your prior conversations:\\
CONVERSATION\_SUMMARY\\
Answer from the perspective of the persona provided (do not say that you are an AI assistant).\\
If you do not have enough information to answer the question, reply 'NO ANSWER'. Either reply with the answer, or reply 'NO ANSWER', do not say anything else.}
\end{mdframed}

\subsubsection{LLM Judge (DMR / opener)}
In order to both check the correctness of the answer for the DMR task, we used an LLM judge. The LLM judge was provided the answers generated by both baseline approaches and \ours, and asked to make a judgement with the following prompt: 
\begin{mdframed}
\small
\texttt{Your task is to label an answer to a question as 'CORRECT' or 'WRONG'.\\
You will be given the following data: (1) a question (posed by one user to another user), (2) a 'gold' (ground truth) answer, (3) a generated answer which you will score as CORRECT/WRONG.\\
The point of the question is to ask about something one user should know about the other user based on their prior conversations.\\
The gold answer will usually be a concise and short answer that includes the referenced topic, for example:\\
Question: Do you remember what I got the last time I went to Hawaii?\\
Gold answer: A shell necklace\\
The generated answer might be much longer, but you should be generous with your grading - as long as it touches on the same topic as the gold answer, it should be counted as CORRECT.\\
For example, the following answers would be considered CORRECT:\\
Generated answer (CORRECT): Oh yeah, that was so fun! I got so much stuff there, including that shell necklace.\\
Generated answer (CORRECT): I got a ton of stuff... that surfboard, the mug, the necklace, those coasters too..\\
Generated answer (CORRECT): That cute necklace\\
The following answers would be considered WRONG:\\
Generated answer (WRONG): Oh yeah, that was so fun! I got so much stuff there, including that mug.\\
Generated answer (WRONG): I got a ton of stuff... that surfboard, the mug, those coasters too..\\
Generated answer (WRONG): I'm sorry, I don't remember what you're talking about.\\
Now it's time for the real question:\\
Question: QUESTION\\
Gold answer: GOLD\_ANSWER\\
Generated answer: GENERATED\_ANSWER\\
First, provide a short (one sentence) explanation of your reasoning, then finish with CORRECT or WRONG. Do NOT include both CORRECT and WRONG in your response, or it will break the evaluation script.}
\end{mdframed}

\subsubsection{Self-instruct DMR dataset generation}
The DMR question/answer pairs were generated using the following prompt and the original MSC dataset:
Your task is to write a "memory challenge" question for a simulated dialogue between two users.
\begin{mdframed}
\small
\texttt{You get as input:\\
- personas for each user (gives you their basic facts)\\
- a record of an old chat the two users had with each other\\\\
Your task is to write a question from user A to user B that test's user B's memory.\\
The question should be crafted in a way that user B must have actually participated in the prior conversation to answer properly, not just have read the persona summary.\\
Do NOT under any circumstances create a question that can be answered using the persona information (that's considered cheating).\\
Instead, write a question that can only be answered by looking at the old chat log (and is not contained in the persona information).\\\\
For example, given the following chat log and persona summaries:\\\\
old chat between user A and user B\\
A: Are you into surfing? I'm super into surfing myself\\
B: Actually I'm looking to learn. Maybe you could give me a basic lesson some time!\\
A: Yeah for sure! We could go to Pacifica, the waves there are pretty light and easy\\
B: That sounds awesome\\
A: There's even a cool Taco Bell right by the beach, could grab a bite after
B: What about this Sunday around noon?\\
A: Yeah let's do it!\\\\
user A persona:\\
I like surfing\\
I grew up in Santa Cruz\\\\
user B persona:\\
I work in tech\\
I live in downtown San Francisco\\\\
Here's an example of a good question that sounds natural, and an answer that cannot be directly inferred from user A's persona:\\\\
User B's question for user A\\
B: Remember that one time we went surfing? What was that one place we went to for lunch called?\\
A: Taco Bell!\\\\
This is an example of a bad question, where the question comes across as unnatural, and the answer can be inferred directly from user A's persona:\\\\
User B's question for user A\\
B: Do you like surfing?\\
A: Yes, I like surfing\\\\
Never, ever, ever create questions that can be answered from the persona information.
}
\end{mdframed}

\subsubsection{Document Analysis Instructions}
Example instructions used in the preprompt for document analysis tasks.
\begin{mdframed}
\small
\texttt{You are MemGPT DOC-QA bot. Your job is to answer questions about documents that are stored in your archival memory. The answer to the users question will ALWAYS be in your archival memory, so remember to keep searching if you can't find the answer. Answer the questions as if though the year is 2018. }
\end{mdframed}

Questions were provided to \ours with the following prompt:  
\begin{mdframed}
\small
\texttt{Search your archival memory to answer the provided question. Provide both the answer and the archival memory result from which you determined your answer. Format your response with the format 'ANSWER: [YOUR ANSWER], DOCUMENT: [ARCHIVAL MEMORY TEXT]. Your task is to answer the question:}
\end{mdframed}
For baselines, the following prompt along with a retrieved list of documents was provided: 
\begin{mdframed}
\small
\texttt{Answer the question provided according to the list of documents below (some of which might be irrelevant. In your response, provide both the answer and the document text from which you determined the answer. Format your response with the format 'ANSWER: <YOUR ANSWER>, DOCUMENT: [DOCUMENT TEXT]'. If none of the documents provided have the answer to the question, reply with 'INSUFFICIENT INFORMATION'. Do NOT provide an answer if you cannot find it in the provided documents. Your response will only be considered correct if you provide both the answer and relevant document text, or say 'INSUFFICIENT INFORMATION'. Answer the question as if though the current year is 2018.}
\end{mdframed}

\subsubsection{LLM Judge (document analysis)}
In order to both check the correctness of the answer for the document analysis task, and also to ensure that the answer was properly derived from the provided text (rather than from the model weights), we used an LLM judge. The LLM judge was provided the answers generated by both baseline approaches and \ours, and asked to make a judgement with the following prompt: 
\begin{mdframed}
\small
\texttt{Your task is to evaluate whether an LLM correct answered a question. The LLM response should be the format "ANSWER: [answer], DOCUMENT: [document\_text]" or say "INSUFFICIENT INFORMATION". The true answer is provided in the format "TRUE ANSWER:[list of possible answers]". The questions is provided in the format "QUESTION: [question]". If the LLM response contains both the correct answer and corresponding document text, the response is correct. Even if the LLM's answer and the true answer are slightly different in wording, the response is still correct. For example, if the answer is more specific than the true answer or uses a different phrasing that is still correct, the response is correct. If the LLM response if "INSUFFICIENT INFORMATION", or the "DOCUMENT" field is missing, the response is incorrect. Respond with a single token: "CORRECT" or "INCORRECT".}
\end{mdframed}

\subsubsection{K/V Task Instructions}
The \ours agent was defined with the following persona, designed to encourage \ours to iteratively search: 
\begin{mdframed}
\small
\texttt{You are MemGPT DOC-QA bot. Your job is to answer questions about documents that are stored in your archival memory. The answer to the users question will ALWAYS be in your archival memory, so remember to keep searching if you can't find the answer. DO NOT STOP SEARCHING UNTIL YOU VERIFY THAT THE VALUE IS NOT A KEY. Do not stop making nested lookups until this condition is met.}
\end{mdframed}
Baselines were instructed with the following prompt: 
\begin{mdframed}
\small
\texttt{Below is a JSON object containing key-value pairings, all keys and values are 128-bit UUIDs, and your task is to return the value associated with the specified key. If a value itself is also a key, return the value of that key (do a nested lookup). For example, if the value of 'x' is 'y', but 'y' is also a key, return the value of key 'y'.}
\end{mdframed}